\documentclass[conference]{IEEEtran}
\IEEEoverridecommandlockouts
\usepackage{cite}
\usepackage{amsmath,amssymb,amsfonts}
\usepackage{algorithmic}
\usepackage{graphicx}
\usepackage{textcomp}
\usepackage{xcolor}
\usepackage{verbatim}
\usepackage{multirow}
\usepackage{multicol}
\usepackage{pgfplotstable}
\usepackage{color}
\usepackage{svg}
\usepackage{xcolor}
\usepackage{caption}
\usepackage{algorithm}
\usepackage{enumitem}
\usepackage{fancyvrb}
\usepackage{fvextra}

\usepackage{syntax}
\usepackage{subcaption}

\usepackage{todonotes}

\def\BibTeX{{\rm B\kern-.05em{\sc i\kern-.025em b}\kern-.08em
    T\kern-.1667em\lower.7ex\hbox{E}\kern-.125emX}}
    
\definecolor{codegreen}{rgb}{0,0.6,0}
\definecolor{codegray}{rgb}{0.5,0.5,0.5}
\definecolor{codepurple}{rgb}{0.58,0,0.82}
\definecolor{backcolour}{rgb}{0.95,0.95,0.92}
\usepackage[justification=centering]{caption}
\usepackage{listings}

\lstdefinestyle{python}{
    language=Python,
    basicstyle=\ttfamily\scriptsize,
    keywordstyle=\color{blue},
    stringstyle=\color{red},
    commentstyle=\color{green},
    morecomment=[l][\color{magenta}]{\#},
    tabsize=2,
    showspaces=false,
    showstringspaces=false,
  frame=single,
  rulecolor=\color{black!30},
  captionpos=b,
  belowcaptionskip=1em,
  aboveskip=1em,
  belowskip=0em,
  breaklines=true,
  xleftmargin=1em,
  xrightmargin=1em,
    numbers=left,
  firstnumber=1,
  numberstyle=\scriptsize,
}

\lstdefinelanguage{sta}{
  morekeywords={entry,formats,prompt,target,from,append,text,__next,__exit},
  sensitive=false,
  morecomment=[l]{\#},
  morestring=[s]{\{}{\}},
  basicstyle=\ttfamily\scriptsize,
  keywordstyle=\bfseries\color{blue},
  stringstyle=\itshape\color{purple},
  commentstyle=\color{gray},
  showstringspaces=false,
  breaklines=true,
  tabsize=2,
  literate=
    {-\ }{{\textbf{\color{red}- }}}2
    {>\ }{{\textbf{\color{green}> }}}2,
  frame=single,
  rulecolor=\color{black!30},
  captionpos=b,
  belowcaptionskip=1em,
  aboveskip=1em,
  belowskip=0em,
  xleftmargin=1em,
  xrightmargin=1em,
    numbers=left,
  firstnumber=1,
  numberstyle=\scriptsize,
}

\usepackage{hyperref}
\hypersetup{
    colorlinks=true,
    linkcolor=blue,
    filecolor=magenta,      
    urlcolor=cyan,
    pdftitle={Structured Thoughts Automaton},
    pdfpagemode=FullScreen,
    }

\usepackage{caption}
\usepackage{setspace}

\usepackage{etoolbox}

\AtBeginEnvironment{quote}{\par\singlespacing\small}
\AtBeginEnvironment{grammar}{\small}

\usepackage{float}

\begin{document}

\title{Structured Thoughts Automaton: First Formalized Execution Model for Auto-Regressive Language Models\\
\thanks{This work is supported by the U.S. Department of Energy, Office of Science, Advanced Scientific Computing Program under contract number DE-AC02-06CH11357 and Award Number DE-SC0021293. Prepared by LLNL under Contract DE-AC52-07NA27344 (LLNL-CONF-849468).}
}

\makeatletter
\newcommand{\linebreakand}{%
  \end{@IEEEauthorhalign}
  \hfill\mbox{}\par
  \mbox{}\hfill\begin{@IEEEauthorhalign}
}
\makeatother    
\author{
\IEEEauthorblockN{Tristan Vanderbruggen$^{1}$}
\and
\IEEEauthorblockN{Chunhua Liao$^{1}$}
\and
\IEEEauthorblockN{Peter Pirkelbauer$^{1}$}
\and
\IEEEauthorblockN{Pei-Hung Lin$^{1}$}
\linebreakand
\IEEEauthorblockA{{$^1$}Lawrence Livermore National Laboratory, Livermore, CA 94550, USA
}
}

\maketitle

\thispagestyle{plain}
\pagestyle{plain}

\begin{abstract}
In recent months, Language Models (LMs) have become a part of daily discourse, with focus on OpenAI and the potential of Artificial General Intelligence (AGI). Furthermore, the leaking of LLama's weights to the public has led to an influx of innovations demonstrating the impressive capabilities of generative LMs. While we believe that AGI is still a distant goal, we recognize the potential of LMs in solving tasks such as searching complex documents, compiling reports with basic analysis, and providing assistance in problem-solving. In this paper, we propose formalizing the execution model of language models. We investigate current execution models, to find that this formalism has received little attention, and present our contribution: the first formalized execution model for LMs. We introduce a new algorithm for sampling the predictions of LMs, which we use to build a reliable and inspectable execution model.
We introduce a low-level language to write ``cognitive program'' for this execution model. We hope to shed light on the need for execution models for LMs and encourage further research in this area.
\end{abstract}

\begin{IEEEkeywords}
Language Models, Programming Languages, Execution Model, Generative AI, Inspectable AI, AI Algorithms
\end{IEEEkeywords}

\subsection*{Preprint Notes}

This paper has been submitted for peer review.
All examples have a working implementation at the time of writing.
We highlighted a few features that are being implemented.
The framework AutoCog is released under Apache 2.0 license at \url{https://github.com/LLNL/AutoCog}.

\section{Introduction}

Language Models (LMs)~\cite{jing2019survey,zhao2023survey} are commonly used to complete prompts, which are text documents that describe some tasks to be performed.
As we make LMs perform increasingly complex tasks, the syntax of these manually crafted prompts have been growing more complicated.
Well crafted prompts can accept a wide range of data (such as user's question and chat history) without deviating from the task.
It is important as we rely on the LM to provide appropriately formatted text such that we can parse it (usually with regular expressions).
The data parsed from the LM response is used to call tools or trigger other prompts.
As the number of components (prompts and tools) in these system grow, it will rapidly become unmanageable.
The introduction of a formalized Execution Model is the first step to establish a real programming environment for LMs.

Our execution model, Structured Thoughts Automaton (STA), specifically targets auto-regressive language models (ARLM).
STA is equipped with a matching low-level language to enable the creation of ``cognitive programs''.
We introduce STA within AutoCog (Automaton \& Cognition), a python framework to build Cognitive Architecture.
AutoCog defines $Cog$, a class of asynchronous callable objects managed by a cognitive architecture ($CogArch$).
STA programs compile to $STA$ a subclass of $Cog$.
AutoCog's $Cogs$ are easily specialized to provide access to tools such as search engines through their APIs.

With AutoCog, we aim at facilitating the design of execution models beyond ARLM. 
Many think~\cite{kaplan2020scaling,hoffmann2022training} that growing the number of parameters in LLM has reached the point of diminishing returns.
Furthermore, next token prediction (NTP) seems inherently limited in its ability to capture semantics.
However, competing ideas such as Joint Embedding Predictive Architecture (JEPA)~\cite{lecun2022path} are more suited for sequence of images at this stage.
We believe that the execution model is a concept that is missing from modern machine-learning.
It might even be the concept needed to bridge the gap between the symbolic and connectionist views of AI.
We are creating one place to implement:
 (1) execution models (specific to the machine-learning architecture),
 (2) programming models (compilable to some execution models),
 (3) symbolic AI algorithms for LM, and
 (4) training of new ML model by transcribing execution traces across execution models.
In this paper, we present the first execution model with its own low-level language.
It does, technically, also constitute the first programming model as we provide an initial library for writing programs.

\section{State-of-the-art}
\label{sec:soa}

\subsection{Large Language Models}

The Large Language Models (LLMs)~\cite{bubeck2023sparks,anil2023palm} that have made the news lately are specifically Auto-Regressive Transformer-based Language Models. LLMs are a feat of engineering where hundreds, if not thousands, of ``tweaks'' enable widely over-parameterized models to converge. The Transformers model architecture was introduced in~\cite{vaswani2017attention}. Generative Pretrained Transformer (GPT)~\cite{radford2018improving} introduced the combination of auto-regressive transformers for language modeling and large-scale pretraining using Next Token Prediction (NTP). However, auto-regressive language models (ARLM) predate artificial neural networks (ANNs).
In essence, a language model assigns some probabilities to sequence of tokens from an alphabet.
Given a sequence of tokens, a causal language model assigns probabilities to continuations of this sequence.
Finally, ARLM predict the next token given a sequence.
Auto-regressive means that to predict the following token, the previously predicted token is added to the end of the input sequence.
The auto-regressive process applied to language models is often referred to as Next Token Prediction (NTP).

The current technology relies on foundational LLMs that cost hundreds of thousands of dollars to train, though the cost is declining fast.
While the models and software to perform this training are extremely complex, the training itself could not be simpler.
 It is NTP applied billions of times, evaluating the error and propagating that error to adjust the models' billions of parameters.

One of the real breakthroughs of the past few months is the realization that LLMs can be fine-tuned for a few hundreds of dollars, and that we have the techniques to run them at the edge.
LLaMa~\cite{touvron2023llama} is a foundational model that was released for research-purpose by MetaAI.
Soon after its release, LLaMa's weights were leaked to the public, leading to a wave of innovation.
Stanford Alpaca~\cite{taori2023stanford} was fine-tuned from LLaMa for less than \$600.
Alpaca-LoRA~\cite{tloen2023alpaca} enables fine-tuning on consumer hardware such as a gaming GPU.
LLaMa.cpp~\cite{ggerganov2022llamacpp} was ``hacked in an evening'' and was soon capable of running LLaMa-based models on Raspberry Pi and Pixel 6.

\subsection{Intrinsic Execution Model}

When we mention Execution Model in the context of ARLM, we mean three things:
  (1) how do we assemble the input sequence,
  (2) how do we generate new tokens, and
  (3) what happens to the generated tokens.

The most common execution model used for LMs is Next Token Prediction (NTP), which is the initial execution model for most Generative LMs. NTP involves predicting the next token in a sequence given the preceding tokens. In some cases, pretraining may have used Masked Language Modeling (MLM\footnote{colloquially known as \emph{fill-the-blanks}: sentence or paragraph with missing words, students must figure-out those missing words.}) with an encoder architecture, such as BERT~\cite{devlin2019bert}, but eventually, it is fine-tuned for NTP when used for generative tasks.

While NTP is not very useful on its own, it is used to implement various completion algorithms. The straight application of NTP is colloquially referred to as \emph{greedy}, but most generative systems use variations of the beam search algorithm which is often referred to as \emph{completion}. \emph{Truncation} is a very simple execution model that builds on \emph{completion}. It deals with preventing termination because the token window is full. It is used to build story-teller and chatbot systems, by truncating from the head or middle, respectively.

\subsection{Special tokens}

Special tokens are another way some form of execution model is enforced. These tokens do not come from the source language but are added to control the LM. Classic examples are start/end of text/document and blank. Modern generative LLMs, often support a small set of special tokens used to organize the instruction. For example, the recently released StarCoder~\cite{li2023starcoder} has the following: \Verb$<|system|>$, \Verb$<|user|>$, \Verb$<|assistant|>$, and \Verb$<|end|>$. The first three start text sections while the last ends those sections. The \Verb$<|system|>$ section comes first, it \emph{adjusts} the purpose of the model. It is followed by a \Verb$<|user|>$ section to specify the input. Finally, StarCoder fills the \Verb$<|assistant|>$ section until it produces the \Verb$<|end|>$ token.

Special tokens are also instrumental to implement training techniques such as MLM where mask tokens are used. In that case, the input sequence is masked at random using special \emph{mask} tokens. Each mask token appears only once in the masked sequence. The previous sentence could become:
\begin{Verbatim}[breaklines=true]
<|input|>Each <|M1|> token appears only <|M3|> in the masked <|M2|>.<|end|> <|answer|><|M1|>mask<|M2|>sequence <|M3|>once<|end|>
\end{Verbatim}
The LM is then given the \Verb$<|input|>$ section and trained to produce the \Verb$<|answer|>$ section.

We find it revealing that special tokens work.  It shows a \emph{willingness} from the model to follow sequences and use variables.
In fact, special tokens form a communication protocol above the natural language.
Furthermore, even smaller LM are very good at ``artificial'' syntax, like python, CSS, HTML, JSON, and Markdown.
It seems to us that LM are particularly good at syntax but have a shallow understanding of semantic.
The focus on syntax over semantic could be inherent to NTP.

Our execution model does not use special tokens to ensure compatibility with current ARLMs. However, it is our goal to eventually separate \emph{data} and \emph{command} tokens.

\subsection{Emerging Execution Model}

LangChain~\cite{langchain2023} is a framework that allows the building of pipelines of prompts. Each stage can iterate between completion and python logic to complete its prompt. LangChain's Agents use completion and regular expression to control external tools, allowing the LM to extract information from data formatted by the agent.
LangChain implements many state-of-the-art agents such as Reasoning/Acting (or ReAct)~\cite{yao2023react}.
ReAct presents the LLM with a prompt that describes a task, a list of tools, and a prompting format. 
The format section explains how the LLM is suppose to (1) think, (2) pick a tool, (3) provide inputs for the tool, (4) observe the output of the tool, and (5) loop back. One of the tool options is to interrupt the loop.
LangChain Agent provides very little control and must be used with heavily fine-tuned models with a low temperature setting (a measure of the model's ``creativity'').

By chaining multiple agents within LangChain, complex behavior can be elicited from the LM. We hypothesize that the transfer of information across contexts is the source of the sparks of Artificial General Intelligence observed with GPT-4~\cite{bubeck2023sparks}. LLMs have mastered the syntax of both human and artificial languages, enabling them to read and write JSON, facilitating the communication of structured data between symbolic and connectionist processes (python programs and LLMs). It is possible to achieve similar results with smaller models if we have a better execution model. Already, fine-tuning LLMs for a specific set of LangChain prompts can provide impressive results. Furthermore, using Low Rank Adaptation (LoRA)~\cite{hu2021lora}, it is possible to perform this fine-tuning for a couple hundred of dollars.

Over the past few months, most big players in the AI industry have been launching their own line of products or tools to leverage LLM chaining.
There is HugginFace's Transformers Agent that defines a natural language API on top of transformer models. The agents can interpret natural language requests from users and use a set of curated tools through HugginFace APIs in various ML-based workflows~\cite{transformers-agents}.
Both Google\footnote{Google I/O event a few days before submission} and OpenAI~\cite{openai-plugins-introduction} have ``plugins'' which are a variety of tools that the agent can use to complete a task. For example, OpenAI plugins connect ChatGPT to third party applications to retrieve real-time information or assist users with actions. An ai-plugin.json file is used to define a plugin's name, description, endpoints, authentication schema and so on. 
In all case, they use very similar techniques to LangChain's Agents.

Recently, Language Model Query Language (LMQL)~\cite{beurer2022prompting} introduced the idea of Language Model Programming (LMP).
They created a small language to describe prompts and provide some degree of reliability.
It seems the underlying system uses Deterministic Finite Automaton to parse the token stream.
LMQL provides more freedom to the user within the prompt than our work where a precise syntax facilitate the creation of programs that span multiple prompts.

Aside from LMQL, none of these systems consider controlling the tokens that are produced by the LM.
We must assume that the big players can rely on heavily fine-tuned the models as they have the compute and data.
Without that fine-tuning, LLMs will not follow directions in a reliable manner.
However, it is desirable to be able to run LM applications with foundation models quantitized to 4 bits.
That can be used to probe these models, or to enable iterative training without labels.
Our execution model can execute programs on any ARLM, we have used it with OpenAI GPT-3 (API), GPT-2 (HugginFace's transformers), and LLaMa 7B (using LLaMa.cpp and 4 bits quantization -- model has 4GB footprint in RAM).

\section{Structured Thoughts Automaton}
\label{sec:sta}

In this section, we introduce Structured Thought Automaton, or STA, which is a formalized Execution Model for LMs.
STA's main concepts are: (1) structured prompts, (2) communication channels and (3) data formats.
These concepts are captured in a low-level language.
While, we refer to STA as an Execution Model given its very low-level of abstractions.
STA is made of an execution model, a language, and a (tiny) library of programs.
Hence, technically, it is a programming model, albeit a burgeoning one.
We simply wish to convey the fact that proper programming languages must be built.

We will present the three main concepts, followed by the language design, details of the execution model, desription of the execution traces, and finally the choice algorithm.
While the choice algorithm might seem out of place, it is the one trick that make STA possible.
Indeed, it let the LM decide which branch of the automaton should be taken when a choice arise.

\begin{figure}[tbp]
  \centering
  \includegraphics[width=.8\linewidth]{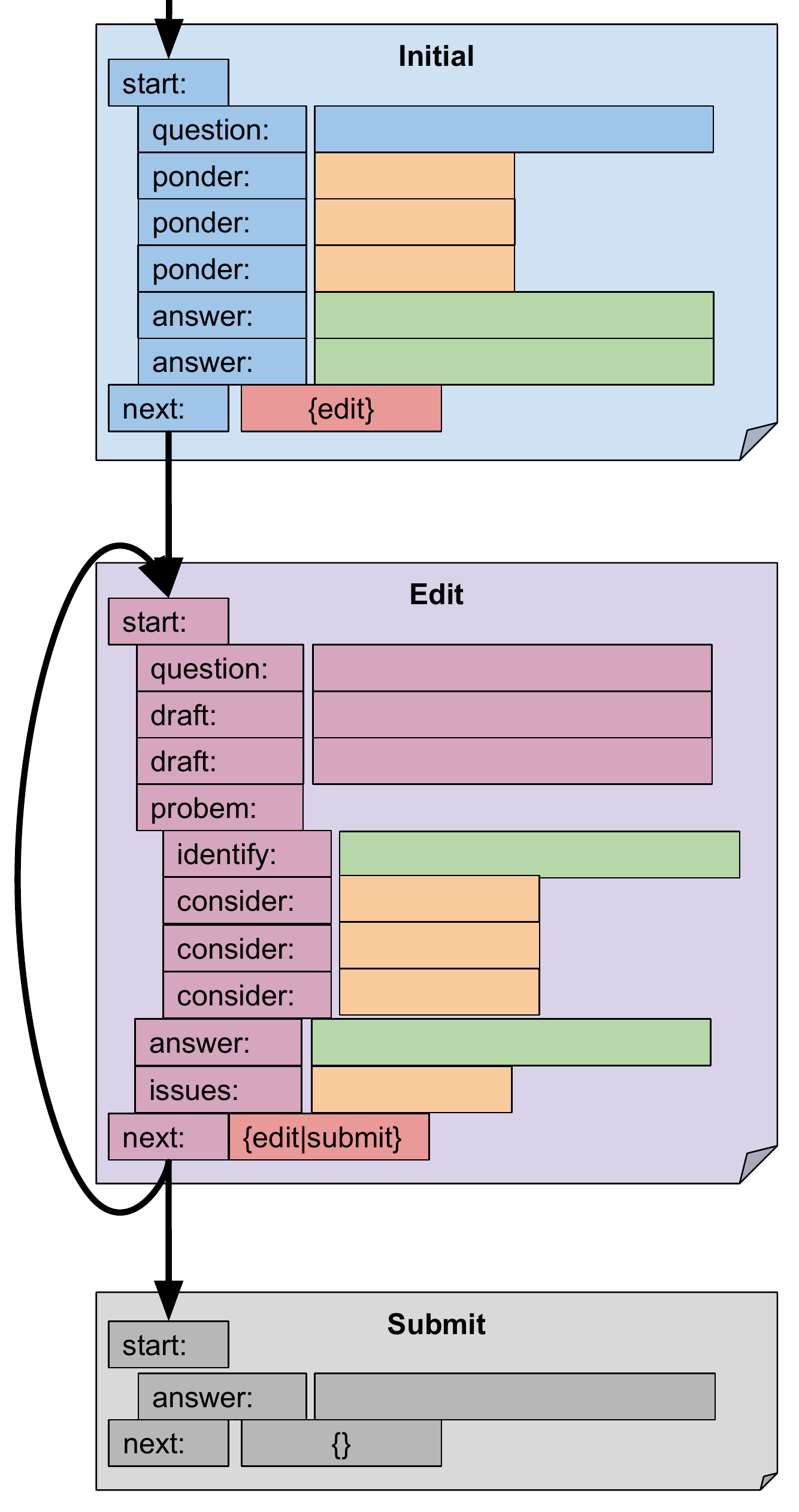}
  \caption{
Illustration of our example program (Figure~\ref{fig:sta:language:motivation}).
The three main boxes represent prompts with the hierarchical questionnaire.
Edges represent the control-flow of the program that is decided by the LM.
Empty line of the same color as the prompts represent inputs (w.r.t. the prompt).
Green and orange empty lines are filled by the LM using sentences or thoughts respectively.
We configure the LM to make thoughts shorter and more creative than sentences.
}
    \label{fig:sta:artist}
\end{figure}


\subsection{Main Concepts}
\label{sec:sta:main}
In Figure~\ref{fig:sta:artist}, we illustrate a program implemented with STA.
It is representation of our main example from Figure~\ref{fig:sta:language:motivation}.
STA oversees the execution of \emph{prompts} which produce structured documents.
The leaves of these documents have prescribed \emph{formats}.
Each prompt declares communication \emph{channels} which are executed before the questionnaire.
The questionnaire compiles to a push-down automaton.
For each state, the LM provides text that follows the prescribed format (using either \emph{completion} or \emph{choice}).
When there is more than one possible branch in the PDA, the \emph{choice} algorithm is used.

\subsubsection{Prompts}

Prompts are executed in sequence and each prompt can have any number of successors.
If more than one successor, the last question is to decide which one is next.

\begin{figure}[htbp]
    \lstset{language=sta}
    \begin{lstlisting}
You are a helpful AI assistant.
Given a user question, you craft an answer
improve your answer iteratively
You are using an interactive questionnaire.
Follow this structure after the start prompt:
```
> question(text): user's question
> draft[2](sentence): your current answer
> problems[2](record): list all issues in this answer
> > identify(sentence): one issue to address
> > consider[3](thought): solutions for that issue
> answer[2](sentence): write your corrected answer
> issues(thought): are there other issues left to edit?
```
Each prompt expects one of the following formats:
- next: "edit" the issues or "submit" your answer
- text: ASCII text in any form
- sentence: one natural language sentence per line
- record: start of a nested prompt
- thought: your thoughts (a few words per lines)
Terminate each prompt with a newline. Use as many statement with "thought" format as needed.

start(record):
> question(text): Explain the different phases of a compiler
> draft[1](sentence):  A compiler typically goes through several distinct phases to translate source code into executable code.
> draft[2](sentence):  These phases include lexical analysis, syntax analysis, semantic analysis, code generation, and optimization.
> problems[1](record):
> > identify(sentence):  This answer is too technical for a general audience.  
> > consider[1](thought):  Omit nonessential details.  
> > consider[2](thought):  Use simpler language.  
> > consider[3](thought):  Add an example of one of the phases.  
> answer[1](sentence):  A compiler translates source code into executable code in several steps, such as analyzing the code, generating the code, and optimizing it. For example, the lexical analysis phase scans the source code to identify the individual symbols it contains.  
> issues(thought):  No, the answer is now complete.
exit(next): submit
    \end{lstlisting}
    \caption{
Transcript of OpenAI's GPT-3.5 running the \Verb{edit} prompt from the program of Fig~\ref{fig:sta:language:motivation}.
In this example, we asked GPT to ``explain the different phases of a compiler''.
Its original answer (shown as \Verb{draft}) was pretty decent but it decided it was too technical.\\
As usual with LLM, tiny changes in any wording can completely change the results, this is conducting to some fun tuning the program.
For example, in a similar case, GPT-3 decided it had to use metaphors to make the answer more accessible.
The resulting story about a chef cutting vegetable had little to do with a compiler...\\
Amusingly, GPT-3 always thought that compilers were either too technical or complicated.
    }
    \label{fig:sta:main:trace}
\end{figure}

A prompt's header (Fig~\ref{fig:sta:main:trace}~L1-L23) has a set of instructions (Fig~\ref{fig:sta:main:trace}~L7-L13) and a description of the text formats (Fig~\ref{fig:sta:main:trace}~L16-L20) used to answer each question.
After the start prompt the prompt's PDA is used to generate the structure, presenting the LM with a choice when needed.
The content of each line can either be provided through a channel (Fig~\ref{fig:sta:main:trace}~L24-L26) or generated by the LM.
STA uses the format associated with each question to properly configure the completion algorithm (effectively selecting the proper LM wrapper in the cognitive architecture).

In this run, the choice algorithm is used at: line 30 and 31 to keep adding considerations, line 32 to not add another problem, and line 33 to only have one answer. On this last case, it did write two sentences in that line while it usually keep to one sentence per line when using the sentence format. It is possible that it decided that it already had two sentences\footnote{Anecdotally, we created a program that let GPT 3.5 think ten time ``\Verb{> think[10](thought): think as much as you'd like}'' but hack the instructions to say five instead. In our dozen or so tries, it never went further than five thoughts following the instruction even when given the choice not to.}. There is one final use of \emph{choice} when selecting the next prompt. It had to choose ``edit'' or ``submit'' and picked the latter. It agrees with its statement that there are no issues with his answer.

That is similar to most prompting of LLM, for example LangChain's Agent or LMQL.
One difference is that they let input data be formatted in the header while in STA the header is static.
The main difference is that STA introduces nesting in the questionnaire and the declaration of lists.
The results of the execution of one prompt is a structure document: nested list and dictionary with text at the leaves.
The questionnaire is parsed using a push-down automaton to produce that document.

Initially, we tried to introduce this structure in prompts for OpenAI GPT-3 using LangChain.
GPT-3 had no problem reading the input data (list of ten search results with title, url, and description).
The problems came when asked to answer with nested questions.
It would follow the format for a few lines but soon start to add random blank lines or even comments.
After those blanks it often hallucinated\footnote{Colloquially, ``out-of-distribution'' answers from LLM are referred to as ``hallucination''. It is not to be confounded with the LM providing non-factual information. Hallucinations are when the model switch to a completely different subject.} new prompts...
Given the results, we realized that we had to read the LM output line by line, properly configuring the LM for each completion.
The next issue was how to decide branches in the PDA.
For example, we want to let the LM write up to ten sentences, how do we decide when it is done?
We started with a greedy algorithm to decide token by token what was the best branch.
Eventually, we devised a proper \emph{choice} algorithm for that task.

\subsubsection{Formats}

With the introduction of the choice algorithm, we were able to better formalize the idea of \emph{format}.
Formats follow a hierarchy with an abstract root that have three children: \emph{text}, \emph{enum}, and \emph{regex}.
The default format is \emph{text} and causes a call to the completion algorithm of the LM.
\emph{Thought} is a child of \emph{text} meant to use a LM configured for short and ``creative'' completion.
This is achieve by setting the number of desired tokens and the ``temperature''\footnote{temperature is a scalar (usually between 0. and 2.) which is used to configure the creativity of the completion algorithm. Depending on the algorithm, other parameters might be used, such as $top_k$ and $top_p$.} of the LM.
The \emph{enum} format uses the \emph{choice} algorithm to decide between a list of tokens.
So far, it is only used for the control-flow between prompts and there is no possibility to declare an \emph{enum} in the language.
Static \emph{enum} are going to be first with a list of choice declared in the program.
The more interesting concept is dynamic \emph{enum} which can take any values from a list (input or previous prompt).
Finally, formats defined by regular-expressions. It requires the development of an appropriate sampling algorithm.
As regular-expressions compile to deterministic finite automaton, we envisage to adapt beam-search to only explore paths that agree with the DFA.
The goal of \emph{regex} formats is to represent integers, floating point numbers, phone number, path, url, ...
Access to dynamic \emph{enum} and \emph{regex} will be essential to truly probe the capabilities of LM.

\subsubsection{Channels}

Each prompt can have many \emph{channels} which are used to move data from the inputs, or previously executed prompts.
Channels can also trigger external calls to any callable component in the architecture.
Data-parallelism can be achieved through the use of mapped channels which create one instance of the prompt for each element of the source list.
A prompt with multiple mapped channels will have has many instances as the cross-product of the sources.

\subsection{Language}
\label{sec:sta:language}
STA's language is primarily a procedural and structured languages, with some declarative features.
It is equipped with a call interface but there is no context sharing.
Calls are issued to callable objects ($Cogs$) in the cognitive architecture:
 other \emph{programs}, vector-stores, or external tools.
STA \emph{programs} have a collections of \emph{prompts}, a set of \emph{formats}, one \emph{entry},
and a few declarative statements (task description using natural language).

\begin{figure}[htbp]
    \centering
    \begin{grammar}
<program> ::= <entry> <formats>? <prompt>*

<entry> ::= "entry(" <identifier> "):" <sentence> <newline>

<formats> ::= "formats:" <format>*

<format> ::= <itemize> <fmt_decl> <sentence> <newline>

<fmt_decl> ::= <identifier> "[" <identifier> "]:" 

<prompt> ::= <header> <channel>* <state>* <leaf>

<header> ::= "prompt(" <identifier> "):" <sentence> <newline>

<channel> ::= <itemize> <flow> <mapped>? <newline>

<flow> ::= (<target>|<append>|<call>) <prompt>? <from>?

<target> ::= "target(" <identifier> ")"

<append> ::= "append(" <identifier> ")"

<call> ::= "call(" <identifier> ")" <kwargs>*

<kwargs> ::= "kwargs(" <identifier> "," <expression> ")"

<expression> ::= <identifier> | <literal>

<prompt> ::= "prompt(" <identifier> ")"

<from> ::= "from(" <identifier> ")"

<state> ::=  <indent>+ <state_decl> <newline>

<state_decl> ::= <identifier> (<count>)? (<fmtmrk>)? ":"

<count> ::= "[" <digit>+ "]"

<fmtmrk> ::= "(" <identifier> ")"

<leaf> ::= "__" ( <next> | <exit> )

<next> ::= "next(" <branch> ( "," <branch> )* ")"

<exit> ::= "exit(" <identifier> ( "," <identifier> )* ")"

<branch> ::= <identifier> ( "[" <digit>+ "]" )?

<indent> ::= "> "

<itemize> ::= "- "

\end{grammar}
    \caption{BNF Grammar of STA's language. We omitted a few trivial rules for brevity (sentence, literal, digit, identifier, and newline).}
    \label{fig:sta:language:grammar}
\end{figure}

In Figure~\ref{fig:sta:language:grammar}, we provide the BNF representation of STA's grammar.
A program consists of an entry, zero or more formats, and prompts. It also have a few declarative statements (Figure~\ref{fig:sta:language:header}) that are not shown in the grammar. These permit users to override any part of the construction of the prompts' headers. This header contains instructions in natural language and a technical description of the prompts mechanics (Fig~\ref{fig:sta:main:trace}~L1-L23).

\begin{figure*}[tbp]
    \centering
    \lstset{language=sta}
    \begin{lstlisting}
# some more control over the final `header` below
prehamble: You are a helpful AI assistant.
postscriptum: Terminate each prompt with a newline. Use as many statement with \"thought\" format as needed.
basics: You are using an interactive questionnaire.
mechs: Follow this structure after the start prompt:
fmts: Each prompt expects one of the following text formats:

# [not-recommended] `header` is used to assemble the actual header of each prompt
#   `automaton` and `prompt` are set with the entry-point and declaration of each prompt, respectively
#   `mechanics` is automatically generated, it is the statements of the prompt but defaults are added
#   `formats` is also generated based on the format used in this specific prompt
header: {prehamble}\n{automaton}\n{prompt}\n{basics}\n{mechs}\n```\n{mechanics}\n```\n{fmts}\n{formats}\n{postscriptum}\n\nstart(record):\n
    \end{lstlisting}
    \caption{
Shows how to configure the generation of the prompt header in STA's language.
    }
    \label{fig:sta:language:header}
\end{figure*}

\subsection{Execution Model}
\label{sec:sta:em}
A basic block is usually defined as a sequence of statements that has a single entry point and a single exit point, and represents a contiguous sequence of instructions that are executed without interruption.  Similarly, \emph{prompts} have single entry point (upon which data communication and calls occur), and a single exit point (either selecting the next \emph{prompts} or exiting with some outputs). Each prompt is a statically-bound \emph{questionnaire} which produces structured documents. Documents are nested lists and dictionaries with native or user-defined (text) \emph{formats} at the leaves. The \emph{questionnaires} of STA's \emph{prompts} compile to push-down automaton (PDA) shown in Figure~\ref{fig:sta:em:pda}.

\begin{figure*}[tbp]
  \centering
  \includegraphics[width=.9\linewidth]{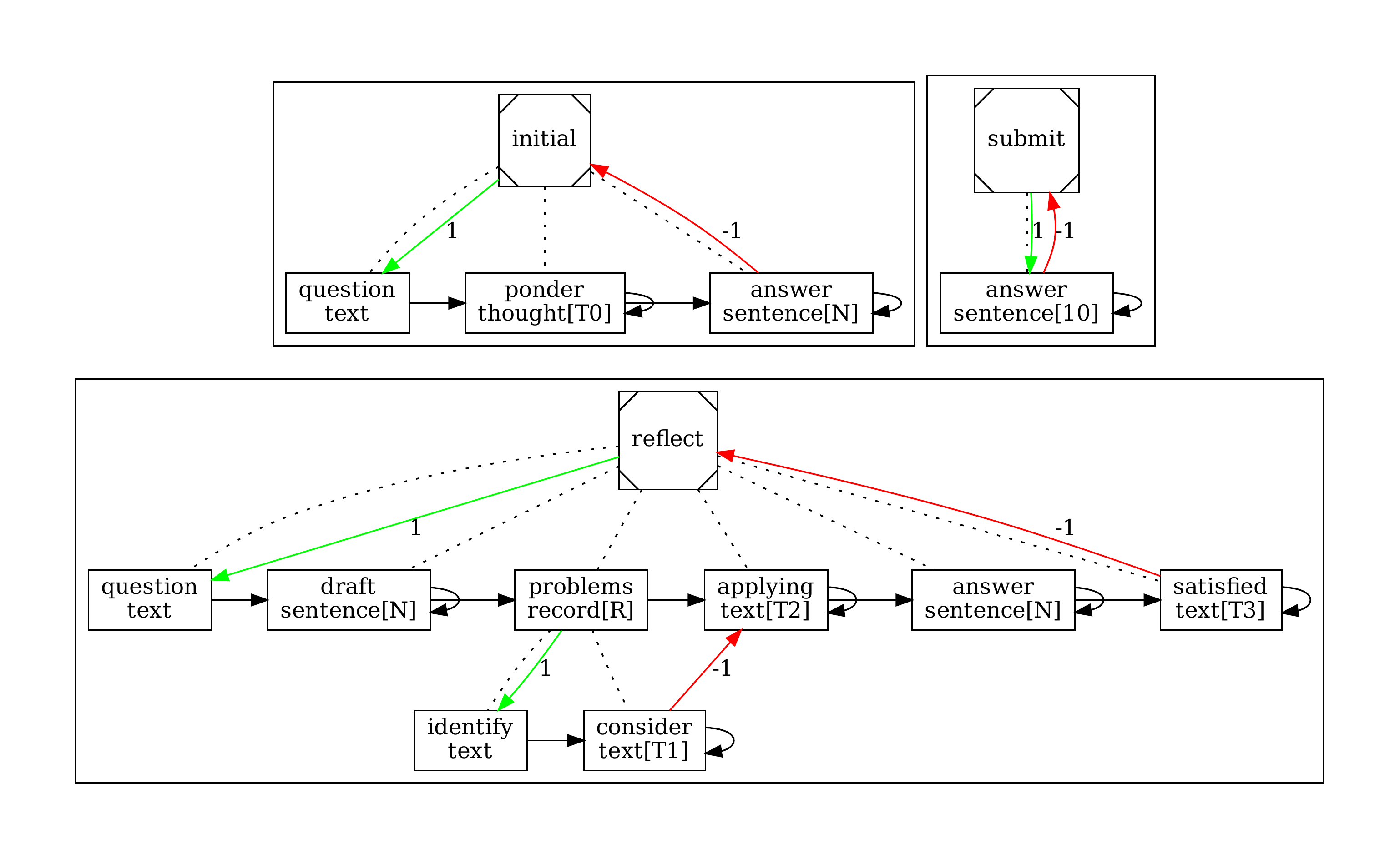}
  \caption{
Graphical representation of the state-machine that implement the Push-Down Automaton of each prompt.
The Hierarchical structure of each prompt is shown with the dotted edges.
The root of each prompt also correspond to the initial state of the PDA.
State transitions are shown with solid edges, with green and red edges corresponding to a push or pop of the stack, respectively.
}
    \label{fig:sta:em:pda}
\end{figure*}

Upon reaching a \emph{prompt}, \emph{channels} are executed first. There are a few types of communication channels: (1) copy from inputs (a) or prompt (b), (2) append, and (3) calls. They can retrieve data from the inputs, the latest \emph{content} of another prompt, or previous \emph{content} of the current prompt.
Channels can be \emph{mapped} causing multiple instantiation of the \emph{questionnaire}. The resulting \emph{questionnaires} are completed independently by the Language Model.

There is one ``soft" constraint which does not influence the current implementation but seems necessary for future efficiency. The organization of the questionnaire should be such that there are no gaps in the communicated data. In this way, after communication, the questionnaire can be loaded with the communicated data and \emph{unparsed} into one contiguous string. Coupled with a standard format to express PDAs (including concepts of \emph{completion} and \emph{choice}), it would enable model vendors to serve API not only for STA but many other execution models compiling to the same PDAs.

Channels of type (1) and (2) are executed first (in declaration order) while type (3) are executed second. This is to simplify the data-flow around calls, particularly for \emph{mapped} channels (only copy and call can be mapped) as they instantiate multiple instances of the questionnaire.

At this time, STA programs are single procedure. We are planning to add alternate entry-points to enable \emph{call} channels to call \emph{functions} that share (some) context with the caller. We are unsure how to deal with sharing context across functions and the parallelism introduced by mapped channels.

\begin{figure}[htbp]
    \centering
    \lstset{language=sta}
    \begin{lstlisting}
entry(initial): Given a user question, you craft an answer

formats:
- sentence(text): one natural language sentence per line

prompt(initial): formulate your initial answer
- target(question)
> question(text): user's question
> ponder[{T}](thought): you think about an answer
> answer[{N}](sentence): your initial answer
__next(edit):

prompt(edit): improve your answer iteratively
- target(question)
- target(draft) source(answer) prompt(initial,edit)
> question(text): user's question
> draft[{N}](sentence): your current answer
> problems[{R}]: list all issues in this answer
> > identify(sentence): one issue to address
> > consider[{S}](thought): solutions for that issue
> answer[{N}](sentence): write your corrected answer
> issues(thought): are there other issues left to edit? 
__next(edit[{L}],submit): "edit" the issues or "submit" your answer

prompt(submit): "ghost" used to join dataflow branches
- target(answer) prompt(edit)
> answer[{N}](sentence):
__exit(answer):
    \end{lstlisting}
    \caption{
This example demonstrates all implemented features of STA (mapped and call channels coming soon behind).
One can note the presence of python f-expression (${x}$), these are \emph{macro} which get substituted before parsing.
In this example, they configure limits on lists' sizes and trip-count of loops.
    }
    \label{fig:sta:language:motivation}
\end{figure}

In Figure~\ref{fig:sta:language:motivation}, we show a sample STA program.
The first line declares that the \Verb{initial} prompt is the entry of this program, it also describe the purpose of this program.
Then, we declare the user-defined \Verb{sentence} format (adding to the native \Verb{text} and \Verb{thought}).
The program uses three prompts to answer a user's question in a few sentences.
\begin{itemize}
    \item The entry prompt, \Verb{initial}, starting at line 6 let the LM produce $T0$ \Verb{thoughts} to ``ponder'' about an initial answer. The answer is made of up to $N$ \Verb{sentences}.
    \item  The second prompt, \Verb{edit}, starting at line 13 takes the current answer and make the LM consider up to $R$ problems in sequence.
Then the LM produces a new version of the answer before deciding whether or not the answer is ready for submission.
    \item After up to $L$ iteration of the second prompt, the final prompt at line 25 is reached.
In this example, it is a ``ghost'' prompt that only serve to join the control-flow before exiting.
\end{itemize}

We originally tried to avoid ``ghost'' prompts but eventually realized that it was causing undue complexity in STA. 
Now that the need for high level programming languages is evident, we embrace ``ghost'' prompts in STA's low level language.

\subsection{Execution Trace}
\label{sec:sta:et}

With an execution model, we can define the notion of execution traces.
In traditional computing, execution traces are obtained thought instrumentation which includes monitoring special hardware registers, introducing special counter in the executable, analyzing snapshots of running processes, and more.
In STA's current implementation, a full trace of the execution is captured.
For a given program, we maintain one stack per prompt.
Each time a prompt is reached a list of \Verb{StructuredThought} objects is stacked (list because of mapped channels).
These objects capture the \emph{content} of the prompt, meaning the document (nested list and dictionary) with the inputs and the LM productions.
Given a program, an input, and the resulting stacks, we can fully reconstruct the execution of the program.

\subsection{Choice Algorithm}
\label{sec:sta:choice}

The \emph{choice} algorithm is a simple concept: ``given a prompt (sequence of token) and a list of candidate completions (list of sequence of tokens) \emph{choose} the most likely completions''.
We have found that perplexity is often used to implement such function.
However, the algorithm shown below is simple, deterministic, and greedily explores all possibilities.
It is consequently expensive to run when dealing with many long candidates.
It could explain the common use of perplexity to compare natural language sentences.
In STA, \emph{choice} is primarily used for branches in the PDA.
These only have a few candidates with shared prefixes.

We use the implementation below with the \emph{LLaMa.cpp} and \emph{HuggingFace} wrappers for LM.
In the case of OpenAI, we do not get access the base \emph{greedy} algorithm (single step prediction with full probability vector) needed to implement \emph{choice}.
It is also worth noting that these implementations are atrociously inefficient, evaluating the full model from scratch for each prediction.

\begin{figure}[htbp]
    \centering
    \lstset{style=python}
    \begin{lstlisting}
class TokenChoiceTree:
    def __init__(self, token=None, depth=0):
        self.token = token
        self.depth = depth
        self.children = {}
        self.proba = None
        self.cumul = 1.

    def __add(self, sequence:List[int]):
        tok = sequence[0]
        if not tok in self.children:
            self.children.update({
              tok : TokenChoiceTree(
                tok, self.depth+1, self)
            })
        tree = self.children[tok]
        return tree if len(sequence) == 1
               else tree.__add(sequence[1:])

    def add(self, llm, text:str):
        return self.__add(llm.tokenize(text))

    def eval(self, llm, prompt):
        if self.token is not None:
            prompt += llm.detokenize([self.token])
        probs = numpy.exp(llm.greedy(prompt))
        for tree in self.children.values():
            tree.proba = probs[tree.token]
            tree.cumul = self.cumul * tree.proba
            tree.eval(prompt)

    def probability(self):
        if self.depth == 0 or self.cumul is None:
            return None
        return numpy.power(self.cumul, 1./self.depth)

    @staticmethod
    def choose(llm, prompt:str, choices:List[str])
        tree = TokenChoiceTree(llm)
        leaves = [ tree.add(c) for c in choices ]
        tree.eval(prompt)
        return numpy.argmax([ l.probability()
                              for l in leaves ]))
    \end{lstlisting}
    \caption{Python Implementation of Choice Algorithm}
    \label{fig:sta:choice:code}
\end{figure}

The rest of this section was generated by ChatGPT (3.5) given the python implementation (Fig.~\ref{fig:sta:choice:code}).
We lightly edited to get the Latex formatting right:
\begin{quote}
The \Verb{TokenChoiceTree} class is a python implementation of the choice algorithm, which is a way to compute the probability of different possible continuations of a given text prompt. The class is initialized with a \Verb{token} (which is \Verb{None} for the root), a \Verb{depth} (which is 0 for the root), and an empty \Verb{children} dictionary. The \Verb{proba} attribute is initialized as \Verb{None} for the root and represents the probability of the current token given its parent. The \Verb{cumul} attribute represents the cumulative probability of the path from the root to the current token, and is initialized as 1.0 for the root. \\

The \Verb{__add} method takes a list of integers (representing tokens) as input and adds the tokens to the tree as children of the current node, returning the last node added. The \Verb{add} method takes a language model (\Verb{llm}) and a string (\Verb{text}) as input, tokenizes the string using the language model, and calls the \Verb{__add} method with the resulting list of tokens (excluding the first token, which is the root). \\

The \Verb{eval} method takes a language model (\Verb{llm}) and a string (\Verb{prompt}) as input. It starts by adding the current token (if it is not the root) to the prompt. It then calls the \Verb{greedy} method of the language model with the updated prompt, which returns the log probabilities of the next tokens. These log probabilities are exponentiated to get the probabilities of the next tokens, which are stored in the \Verb{proba} attribute of each child node. The \Verb{cumul} attribute of each child node is then updated as the product of its parent's \Verb{cumul} and its own \Verb{proba}. Finally, the \Verb{eval} method is recursively called on each child node. \\

The \Verb{probability} method returns the probability of the path from the root to the current node. This is computed as the \Verb{cumul} attribute raised to the power of \Verb{1/depth}, where \Verb{depth} is the depth of the current node in the tree. The \Verb{probability} method returns \Verb{None} for the root and for nodes whose \Verb{cumul} attribute is \Verb{None}. \\

The \emph{choose} method creates a \Verb{TokenChoiceTree} object with a language model (\Verb{llm}) as input. It then adds each text in a list of \emph{choices} to the tree using the \Verb{add} method, and stores the resulting leaves in the \Verb{leaves} list. Finally, it calls the \Verb{eval} method on the root of the tree with a prompt (\Verb{prompt}) and computes the probability of each leaf node using the \Verb{probability} method. The index of the leaf node with the highest probability is returned as the choice.
\end{quote}

\section{Future Work}
\label{sec:future}

Our immediate goal is to facilitate the implementation of symbolic AI algorithms to be executed by connectionist language models.
Symbolic and connectionist visions of AI have been at odd for a few decades.
To bridge the gap between these visions, we will need expressive programming languages.
Particularly, STA provides \emph{formats} which are organized in documents inside a \emph{prompt}.
But it is lacking a notion to represent sub-trees of that document.
These sub-trees, maybe \emph{struct}, would simplify dataflow manipulation (eliminating some ghost prompts) while permiting some code reuse across prompts.
The resulting \Verb{STA+} will also refine the syntax of the language which was cobbled together from the prompting syntax.

Leveraging the type system of \Verb{STA+}, we will introduce State Full Typed Language Model (SFTLM).
Training SFTLM to execute \Verb{STA+} programs enable us to devise new training paradigms.
Particularly, we can use a combination of NTP for state-transitions, JEPA for type-system embeddings, and MLM to generate data tokens for each state.
With formalized execution model, we can leverage LLM to produce training data for SFTLM.
For example, we can run a \Verb{STA+} program with a LLM and save the execution trace.
The resulting traces can then be systematically transcribed to train a SFTLM.
That process is an advanced version of distillation \cite{hsieh2023distilling} where results of running a tuned model is used to train or finetune another model.

In fact, we will investigate grammar-based fine-tuning and self-distillation as ways to train ARLMs.
Fine-tuning is usually done by training for NTP on the whole prompt, with either ground-truth or distilled outputs.
It seems that instead we should focus on the tokens that are expected to be produced by the model.
Furthermore, we will investigate crafting loss functions\footnote{mathematical expression of a model's training objectives} that leverage syntax and type information.

With this grammar-based fine-tuning will investigate self-distillation, an iterative
process to train/tune foundation model with little data.
By design, STA cannot ``crash'' if the LM does not understand, it simply get random output.
The idea of self-distillation is to use increasingly complex curriculum to train the model.
Each curriculum is made of one program, some examples, many exercises, and a grading tool.
For each ``epoch'', the LM run all the exercises which get graded.
Best exercises and examples are used for grammar fine tuning.
We will start with elementary school tasks: spelling, arithmetic, conjugation, ...
Eventually, the curriculum might include complex planning or inductive logic programs.

\section{Conclusion}
\label{sec:conclusion}

We have presented Structured Thoughts Automaton (STA), the first execution model for auto-regressive language models (ARLM).
STA is designed to leverage current Large Language Models by providing a fine level of control over the algorithms used to sample tokens from the language models.
We have shown how STA can be used to build ``cognitive programs'' using a proto-language.
Cognitive programs are made of prompts organized in a control-flow graph.
Each prompt compiles to a push-down automaton (PDA) and declare communication channels.
The LM uses the \emph{choice} algorithm to ``traverse'' the PDA.
The resulting token stream can be parsed into a structured document.

The concept of execution model could bridge the gap between symbolic and connectionist views of AI.
First, we can now implement algorithms such as planning (forward, backward, ...) on top of the LM.
Second, investigating the models understanding of formal logic within a completely formalized framework becomes possible.
Third, syntax and types within the token stream can be used to create loss functions coupling the symbolic constructs and the connectionist optimization objective.
Finally, one can imagine a ``cognitive compiler'' which given a ``problem'', write the ``cognitive program'' to solve it.
It raises intriguing questions about Turing Completeness.

\bibliographystyle{IEEEtran}
\bibliography{references}

\section*{Appendix: LLaMa Traces}

One of the key trigger for this project was the possibility of running LM at the edge with \Verb{LLaMa.cpp}.
We rapidly realized that such model could not drive LangChain's prompts.
While the project has mostly been developed and tested against OpenAI's GPT-3.5, we have made sure that the original goal still stood.
As expected, it can perfectly follow the syntax and even does better than we thought.

Figure~\ref{fig:sta:main:trace-llama} shows a trace running a very similar program as above with LLaMa 7B quantitized at 4 bits (4GB RAM).
The main difference is that it had more space in the list and we were tuning the name of the last question.
That particular version of the last question was confusing as the statement ``ready (to submit)'' was associated with ``are there other issues left to edit?'': this two statements are antithetic.

The small confusion did not matter for this LLaMa who barely understand the instructions.
It does however produces a somewhat logical sequence of statements without crashing the parser.
It means that we can actually build applications without fearing it crashing because the LM replied something unexpected.

\begin{figure}[tbp]
    \centering
    \lstset{language=sta}
    \begin{lstlisting}
You are a helpful AI assistant.
Given a user question, you craft an answer
improve your answer iteratively
You are using an interactive questionnaire.
Follow this structure after the start prompt:
```
> question(text): user's question
> draft[5](sentence): your current answer
> problems[3](record): list all issues in this answer
> > identify(sentence): one issue to address
> > consider[3](thought): solutions for that issue
> answer[5](sentence): write your corrected answer
> ready(thought): are there other issues left to edit?
```
Each prompt expects one of the following formats:
- next: another round of "edit" or ready to "submit"
- text: ASCII text in any form
- sentence: a natural language sentence
- record: start of a nested prompt
- thought: your thoughts (a few words per lines)
Terminate each prompt with a newline. Use as many statement with "thought" format as needed.

start(record):
> question(text): Explain the different phases of a compiler
> draft[1](sentence):  Okay, here is my initial answer:
> draft[2](sentence):  You are right. The compiler goes through different phases in order to create an executable from the source code
> draft[3](sentence):  First of all it needs a parse phase and then the syntax phase
> draft[4](sentence):  It then does a semantical check, which makes sure that everything is legal in your program
> draft[5](sentence):  Then, it performs an optimization to make your code faster. This stage of compilation depends on how you
> problems[1](record):
> > identify(sentence):  the semantic phase
> > consider[1](thought):  The semantic checks ensure that everything is legal in your program
> > consider[2](thought):  It is checking if there are any syntax errors or warnings in your code.
> > consider[3](thought):  Syntax checking will only catch things like missing parenthesis, brackets etc. but it can't
> answer[1](sentence):  The semantic phase of the compiler is to ensure that every thing is legal in your program
> answer[2](sentence):  You are right. Now that we have covered all the phases of a compiler let us take a look
> answer[3](sentence):  at an example. Imagine you want to build a house and you can't figure out how
> answer[4](sentence):  The builders will first need to understand what it is that you want. They will start by doing
> answer[5](sentence):  the analysis phase. In this step they will take into account your budget, number of people and
> ready(thought):  Now that we have covered all the phases of a compiler let us take a look at an example to
exit(next): submit
    \end{lstlisting}
    \caption{
Transcript of LLaMa 7B (4-bits quantization) running a STA prompt.
This is the smallest LLaMa possible, it only need 4GB of RAM.
    }
    \label{fig:sta:main:trace-llama}
\end{figure}

\begin{figure}[htbp]
    \centering
    \lstset{style=python}
    \begin{lstlisting}
import os, sys, json
from autocog import CogArch
from autocog.lm import Llama, OpenAI
from autocog.architecture import PromptTee

arch = CogArch(cogctx={ 'prompt_out':PromptTee(prefix='cgo24', tee=sys.stdout, fmt='{p}/{c}/{t}-{i}.txt')})
sta_fortune = arch.load(tag='cgo24', filepath='./library/cgo24.sta', T=3, N=2, R=2, S=3, L=2)

arch.cogs['cgo24'].LMs.update({
  'text'     : OpenAI(max_tokens=20, temperature=0.4),
  'thought'  : OpenAI(max_tokens=15, temperature=1.0),
  'sentence' : OpenAI(max_tokens=50, temperature=0.7),
})
res = await arch('cgo24', question="Explain the different phases of a compiler")
print(json.dumps(res[0], indent=4))

arch.cogs['cgo24'].LMs.update({
  'text'     : Llama(model_path="7B/ggml-model-q4_0.bin", n_ctx=2048, defaults={'max_tokens':20}),
  'thought'  : Llama(model_path="7B/ggml-model-q4_0.bin", n_ctx=2048, defaults={'max_tokens':15}),
  'sentence' : Llama(model_path="7B/ggml-model-q4_0.bin", n_ctx=2048, defaults={'max_tokens':50}),
})
res = await arch('cgo24', question="Explain the different phases of a compiler")
print(json.dumps(res[0], indent=4))
    \end{lstlisting}
    \caption{
Python code to use an STA program with either OpenAI (GPT 3.5 by default) or LLaMa.cpp.
Notice the \Verb{await}, this code runs in jupiter-lab (where the \Verb{asyncio} loop is already running).
    }
    \label{fig:sta:main:usage}
\end{figure}

\end{document}